\documentclass[letterpaper, 10 pt, conference]{ieeeconf}  

\IEEEoverridecommandlockouts                              

\usepackage{graphics} 
\usepackage{epsfig} 
\usepackage{mathptmx} 
\usepackage{times} 
\usepackage{amsmath} 
\usepackage{amssymb}  

\newcommand{\etal}{\textit{et al.}}
\newcommand{\eg}{\textit{e.g.,}}
\newcommand{\ie}{\textit{i.e.,}}
\newcommand{\etc}{\textit{etc}.}

\usepackage{booktabs}
\usepackage{subcaption} 
\let\labelindent\relax
\usepackage{enumitem}
\usepackage[noadjust]{cite}

\usepackage{hyperref}
\usepackage{xcolor}

\title{\LARGE \bf
Exploring Dynamic Context for Multi-path Trajectory Prediction
}

\author{Hao Cheng$^{1,*}$, Wentong Liao$^{2,*}$, Xuejiao Tang$^{2}$, Michael Ying Yang$^{3}$, Monika Sester$^{1}$, and Bodo Rosenhahn$^{2}$
\thanks{$^*$Equal contribution, name in alphabet order}
\thanks{$^{1}$Institute of Cartography and Geoinformatics, Leibniz University Hannover, Germany,
        {\tt\small \{cheng, sester\}@ikg.uni-hannover.de}}%
\thanks{$^{2}$Institute of Information Processing, Leibniz University Hannover, Germany,
        {\tt\small \{lastname\}@tnt.uni-hannover.de}}%
\thanks{$^{3}$Scene Understanding Group, University of Twente, The Netherlands,
        {\tt\small michael.yang@utwente.nl}}%
\thanks{This work is supported by the German Research Foundation (DFG) through the Research Training Group SocialCars (GRK 1931) and Germany’s Excellence Strategy within the Cluster of Excellence PhoenixD (EXC 2122).}
}

\begin{document}
\maketitle
\thispagestyle{empty}
\pagestyle{empty}

\begin{abstract}
To accurately predict future positions of different agents in traffic scenarios is crucial for safely deploying intelligent autonomous systems in the real-world environment. 
However, it remains a challenge due to the behavior of a target agent being affected by other agents dynamically and  there being more than one socially possible paths the agent could take.
In this paper, we propose a novel framework, named Dynamic Context Encoder Network (DCENet).
In our framework, first, the spatial context between agents is explored by using self-attention architectures. Then, the two-stream encoders are trained to learn temporal context between steps by taking the respective observed trajectories and the extracted dynamic spatial context as input.
The spatial-temporal context is encoded into a latent space using a Conditional Variational Auto-Encoder (CVAE) module.
Finally, a set of future trajectories for each agent is predicted conditioned on the learned spatial-temporal context by sampling from the latent space, repeatedly.
DCENet is evaluated on one of the most popular challenging benchmarks for trajectory forecasting \textit{Trajnet} and reports a new state-of-the-art performance. It also demonstrates superior performance evaluated on the benchmark \textit{inD} for mixed traffic at intersections. A series of ablation studies is conducted to validate the effectiveness of each proposed module.
Our code is available at \url{https://github.com/wtliao/DCENet}.
\end{abstract}

\section{Introduction}
\label{sec:intro}
\begin{figure}[t]
\begin{center}
 \includegraphics[clip=true, trim=0pt 0pt 0pt 0pt, width=0.8\linewidth]{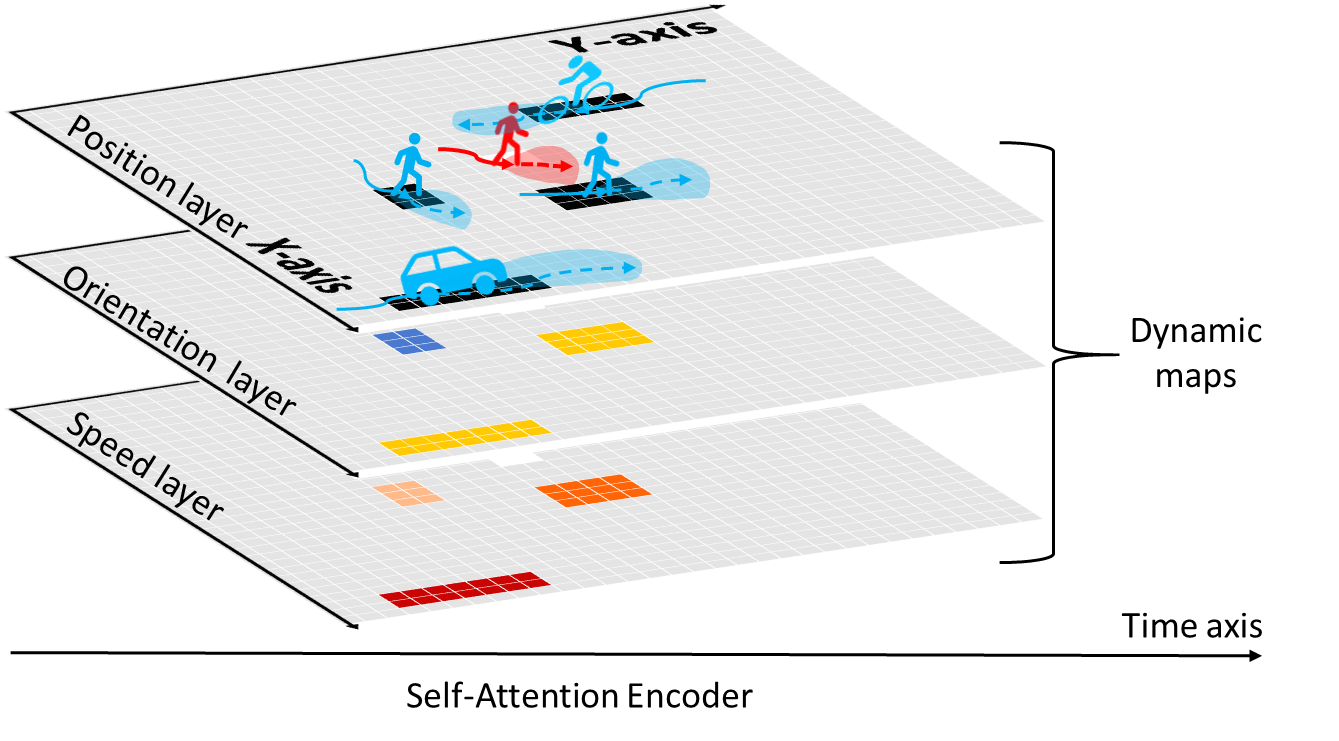}
\end{center}
   \caption{Predicting multiple future trajectories (the most-likely one indicated by dash line over multiple ones indicated by shadow area) of a target agent (in red) conditioned on its observed movement (solid line) with the consideration of its interactions between neighboring agents (in blue) in mixed traffic. Interaction is learned through a sequence of dynamic maps at each step over the time axis and three layers are dedicated to capturing position, orientation and speed information (indicated by color-coded rectangles) using the self-attention structure.}
\label{fig:example}
\end{figure}

Intelligent autonomous systems, such as robots and autonomous vehicles, have a high demand for the ability to accurately perceive, understand and predict the future behavior of humans for effective and safe deployments in our real-world environment.
For example, an autonomous agent will adjust its moving path according to the possible locations of other agents to prevent obstructions or collisions.
However, it is challenging to predict the future location of an agent because it is not deterministic: (1) an agent may change its mind during the movement, (2) other agents' behaviors will affect its next step (\eg~to avoid collisions), and (3) the influence from other agents is dynamic.
Therefore, it is more beneficial to predict a set of potential trajectories adaptive to the dynamic interactions between agents than to predict a deterministic one.
In this work, we seek to explore the dynamic context between agents in traffic scenarios to predict multiple possible trajectories for each agent in the short future (12 steps) by observing their trajectories (8 steps), as showcased in Fig~\ref{fig:example}.

\begin{figure*}[t!]
	\centering
	\includegraphics[clip=true,trim=0pt 0pt 0pt 0pt, width=0.95\linewidth]{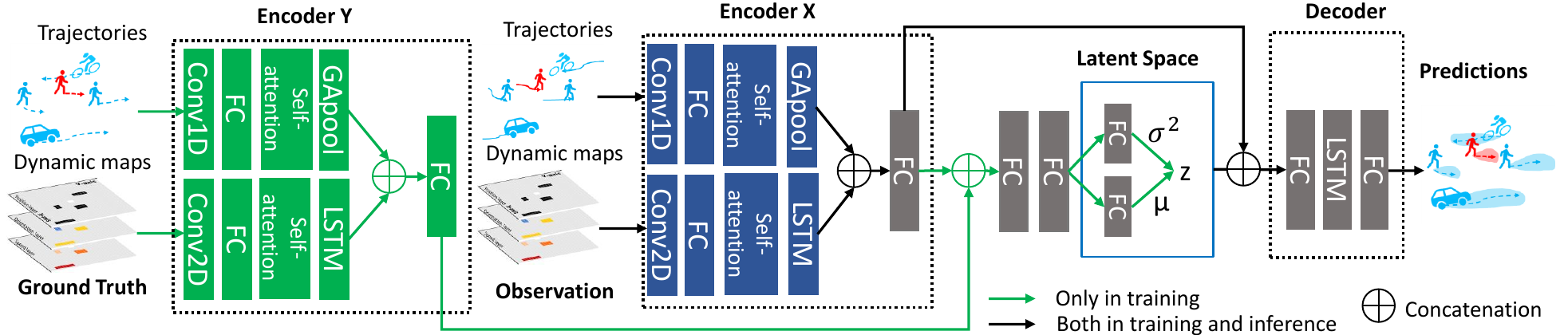}
	\caption{The pipeline for the proposed method. 
    The Encoder Y and Encoder X are identical in structure.}
	\label{fig:pipeline}
\end{figure*}

Specifically, the main contributions of this work are as follows:
(1) It provides a novel framework to predict trajectories of heterogeneous agents (pedestrians, bicycles, vehicles, \etc) in various traffic situations, \ie~20 different shared spaces and four intersections with mixed traffic.
(2) Self-attention modules are integrated into our framework to explore the dynamic context among agents.
(3) A set of possible trajectories for each agent is predicted conditioned on its observed trajectory and the learned dynamic context using a CVAE~\cite{kingma2014semi,sohn2015learning} module.
Extensive experiments are conducted on two of the most popular benchmarks Trajnet challenge~\cite{sadeghiankosaraju2018trajnet} and the new large-scale benchmark inD~\cite{inDdataset} to validate the effectiveness of DCENet for trajectory forecasting.
To judge the effectiveness of each proposed module, we conduct additional ablation studies. 
An overview of our framework is depicted in Fig.~\ref{fig:pipeline}.

\section{Related Work}
\label{sec:related}

\textbf{Trajectory Prediction.}
Forecasting human trajectory has been researched for decades.
In the early stages, many classic approaches are widely applied such as linear regression and Kalman filter~\cite{harvey1990forecasting}, Gaussian processes~\cite{tay2008modelling} and Markov decision processing~\cite{makris2002spatial,kitani2012activity}. 
These traditional methods heavily rely on the quality of manually designed features, which cannot work reliably in a real-world environment of complex spatial-temporal dynamics and are poor at scaling up for dealing with a large amount of data.  
In recent years, many artificial intelligent (AI) technologies have been boosted by the cutting-edge deep learning technologies~\cite{lecun2015deep}, including human trajectory prediction \cite{alahi2016social,gupta2018social,sadeghian2018sophie,zhang2019sr,mohajerin2019multi,tang2019multiple,chandra2019traphic}.
The deep learning models, especially Recurrent Neural Networks (RNNs) with Long Short-Term Memories (LSTMs), show great power in modeling complex social interactions between agents for collision avoidance and exploiting the time dependency for predicting futures~\cite{kothari2020human}.
The Social LSTM network~\cite{alahi2016social} explores the interactions between pedestrians by connecting neighboring LSTMs in the social pooling layer and predicts trajectories for multiple pedestrians.
Zhang~\etal~\cite{zhang2019sr} propose the States Refinement LSTM (SR-LSTM) model that aligns all the agents together and refines the state of each agent through a message-passing framework.
Chandra~\etal~\cite{chandra2019forecasting} combine LSTM and Convolutional Neural Network (CNN) to model the interactions between heterogeneous road agents. 
However, many works have figured out the limited capability of LSTMs in modeling human-human interactions~\cite{becker2018evaluation,becker2018red}. Hence, the attention module~\cite{bahdanau2014neural} is incorporated in LSTMs to learn the spatial-temporal context of trajectories between pedestrians in~\cite{sadeghian2018sophie,al2018move,vemula2018social}.
Recently, the Transformer structure \cite{vaswani2017attention} has shown its power in context learning and sequential prediction
\cite{devlin2019bert,he2020image}. 
In this paper, we will adopt the self-attention module to encode the dynamic interactions between agents.
The recent work~\cite{giuliari2020transformer} seeks to utilize the Transformer structure to predict trajectory instead of LSTMs.
Our work is different from it essentially: (1) we use the generic self-attention module rather than the Deep Bidirectional Transformers (BERT)~\cite{devlin2019bert}, which is a heavy stacked Transformer structure and is pre-trained on large-scale datasets, and (2) our framework is a generative model.

\textbf{Multi-path Trajectory Prediction.}
Many approaches have been proposed to predict a socially compliant set of possible trajectories for an agent~\cite{lee2017desire,gupta2018social,amirian2019social,makansi2019overcoming,poibrenski2020m2p3,cui2019multimodal,katyal2020intent}.
Generative Adversarial Nets (GAN)~\cite{goodfellow2014generative} and CVAE~\cite{kingma2014semi,sohn2015learning} are the most popular generative models used for this task.
In~\cite{gupta2018social} a trajectory sampler named  Social GAN is proposed that considers the social effects of all agents. The generator is trained to predict a set of trajectories for each agent against a recurrent discriminator.
In~\cite{sadeghian2018sophie} social and physical attention mechanisms are implemented in the GAN sampler to predict paths for each agent.
In~\cite{lee2017desire}, multiple plausible prediction samples are generated by a CVAE-based RNN encoder-decoder conditioned on observations.
Katyal~\etal~\cite{katyal2020intent} propose to predict the intent of the target agent using a Bayesian approach as a condition of their CVAE-based LSTM encoder-decoder to help generate multiple paths. Meanwhile, they introduce an LSTM discriminator to train the framework in an adversarial way.
Salzmann~\etal~\cite{Salzmann20Trajectron} propose a CVAE-based model using spatial-temporal graphs to predict pedestrian and car trajectories.  
In \cite{cheng2020mcenet}, scene context and the interactions between individual and group agents are accounted as a condition in a CVAE-based framework to sample multiple trajectories.
\cite{Yuan20Diverse} applies a determinantal point process to increase the diversity sampling of a CVAE-based model for 2D and 3D motion prediction using synthetic data. 
Some other works treat the multi-path trajectory prediction problem as the estimation of a multimodal distribution.
Cui~\etal~\cite{cui2019multimodal} propose to model the multimodality of vehicle movement prediction with Deep Convolutional Networks.
In \cite{makansi2019overcoming}, first, the multimodal distributions are predicted with an evolving strategy by combining the Winner-Takes-ALL loss~\cite{guzman2012multiple}. Then, the samples from the first stage fit a distribution for trajectory prediction. 
Cheng~\etal~\cite{cheng2021amenet} propose AMENet that only employs the self-attention mechanism~\cite{vaswani2017attention} for learning agent-to-agent interaction. In comparison, DCENet adopts a two-stream architecture~\cite{simonyan2014two,casas2018intentnet} of attention modules, with respective streams dedicated to learning the spatial and temporal contexts explicitly.

\section{Method}
\label{sub:method}

\subsection{Problem Formulation}
\label{subsec:problemformulation}

Trajectory prediction is defined as to sequentially predict the future positions $\hat{\mathbf{Y}}_i=\{\hat{\mathbf{y}}_i^{T+1},\cdots,\hat{\mathbf{y}}_i^{T'}\}$ of target agent $i$ by observing its trajectory $\mathbf{X}_i=\{\mathbf{x}_i^1,\cdots,\mathbf{x}_i^{T}\}$, where $\mathbf{x}_i^t = (x_i^t, y_i^t)$ is the coordinates at the $t$-th step and $1\leq t\leq T$. Similarly, $\hat{\mathbf{y}}_i^{t'} = (x_i^{t'}, y_i^{t'})$ is the coordinates at the $t'$-th step and $T<t'\leq T'$. $T$ is the length of observed trajectory and $T'$ is the total length of being observed and predicted trajectory in discrete time steps. $\hat{\mathbf{Y}}_i$ should be as close to the corresponding ground truth $\mathbf{Y}_i$ as possible.
The problem of multi-path trajectory prediction can be formulated as predicting a set of trajectories $\hat{\mathbf{Y}}_i = \{\hat{\mathbf{Y}}_{i,1},\cdots,\hat{\mathbf{Y}}_{i,N}\}$ by observing $\mathbf{X}_i$ for agent $i$, where $N$ is the total number of predicted trajectories.

\subsection{Dynamic Maps}
\label{subsec:maps}
To model the interactions among agents, we first create dynamic maps for each agent that consist of the orientation, speed and position layers of its intermediate environment. These dynamic maps are different from the ones in~\cite{casas2018intentnet} that are designed for modeling map rasterization and traffic lights.  
Centralized on the target agent, a map is defined as a rectangular area of size $W \times H$ and divided into grid cells.
First, referring to the target agent $i$, the neighboring agents $\mathsf{N}(i)$ are mapped into the closest grid $\text{cells}_{w \times h}^t$ according to their relative position as well as the cells reached by their anticipated relative offset (speed) in the $x$ and $y$ directions:
\begin{equation}
\label{eq:cell}
\begin{split}
  &\text{cells}_{w}^t = x_j^t-x_i^t + (\Delta x_j^t - \Delta{x}_i^t), \\
  &\text{cells}_{h}^t  = y_j^t-y_i^t + (\Delta y_j^t - \Delta{y}_i^t), 
\end{split}
\end{equation}
where $w \leq W ,~ h \leq H,~ j \in \mathsf{N}(i)~\text{and}~ j\neq i$.
The \emph{orientation layer} $O$ stores the heading direction that is defined as the angle $\vartheta_j$ in the Euclidean plane and calculated in the given radians by $\vartheta_j = \text{arctan2} (\Delta y_j^t,\, \Delta x_j^t)$. $(\Delta y_j^t,\, \Delta x_j^t)$ is the offset of the position from \textit{t}-th step to the next one for neighboring agent $j$. The angle is shifted into degree $[0,\,360)$. 
Similarly, the \emph{speed layer} $S$ stores the travel speed and the \emph{position} layer $P$ stores the position using a binary flag in the cells mapped above. Last, layer-wise, a Min-Max normalization scheme is applied for normalization, see Fig.~\ref{fig:example}. The map should cover a large vicinity area. Empirically we found $32\times32\,m^2$ a proper setting considering both the coverage and the computational cost. The cell size is set to $1\times1\,m^2$ as a balance to avoid the overlap of multiple agents in one cell based on the distribution of the experimental data, which is also supported by the preservation of personal space~\cite{gerin2005negotiation}.  

\subsection{Encoder Network}
\label{subsec:encoder}
The spatial-temporal context from both the observation time and prediction time are encoded by Encoder X and Y, respectively.  
Both encoders have the same two-stream structure: both streams consist of stacked self-attention layers; as illustrated in Fig.~\ref{fig:pipeline} one stream is followed by a global average pooling (GApool), while the other one is followed by an LSTM module. 
The upper stream is trained to learn motion information from the observed trajectory, whose input is the locations vector of the observed trajectory of the target agent $\mathbf{X}_i=\{\mathbf{x}_i^1,\cdots,\mathbf{x}_i^{T}\}\in \mathbb{R}^{T\times2}$. The lower stream is trained to explore dynamic interactions among agents from the dynamic maps noted as $DM=\{O, S, P\}\in \mathbb{R}^{T\times H \times W \times 3}$ (discussed in Sec.~\ref{subsec:maps}).
For simplicity, we take the upper stream for illustration.
To get a sparse high dimensional representation, $\mathbf{X}_i$ is first passed to a 1D convolution layer (Conv1D) and a fully connected (FC) layer. Each of them is followed by a ReLU non-linear activation. We denote this operation as $\pi(\mathbf{X}_i)$.
A self-attention layer takes as input the Query ($Q$), Key ($K$) and Value ($V$) and outputs a weighted sum of the value vectors.
The weight assigned to each value is calculated as the dot-product of the query with the corresponding key:
\begin{equation}
\label{eq:attention}
    \text{Attention}(Q, K, V) = \text{softmax}(\frac{QK^\mathbf{T}}{\sqrt{d_k}})V,
\end{equation}
where $\sqrt{d_k}$ is the scaling factor, $d_k$ is the dimension of the vector $K$ and $\mathbf{T}$ is the transpose operation. This operation is also called \emph{scaled dot-product attention}~\cite{vaswani2017attention}.
The $Q$, $K$ and $V$ are obtained by three separated linear transformations:
\begin{equation}
\label{eq:qkv}
Q = \pi(\mathbf{X})W_Q,~~K = \pi(\mathbf{X})W_K,~~V = \pi(\mathbf{X})W_V,\\
\end{equation} 
where $W_Q, W_K, W_V\in \mathbb{R}^{d_{\pi}\times d_k}$ are the trainable parameters and $d_{\pi}$ is the dimension of $\pi(\mathbf{X})$.



Because the self-attention module takes all inputs at the same time, position encodings are added to the $Q$, $K$ and $V$ at the bottom of each self-attention layer to encode the temporal information. The sine and cosine functions of different frequencies (varying in time here) are the most widely used:
\begin{equation}
\label{eq:position}
\mathbf{p}^t=\{p_{t,d}\}^D_{d=1},~
p_{t,d}=\left\{
            \begin{array}{lr}
                 sin(\frac{t}{10000^{d/D}}),&\text{for}~d~\text{even};  \\
                 cos(\frac{t}{10000^{d/D}}),&\text{for}~d~\text{odd}, 
            \end{array}
        \right.
\end{equation}
where $D=d_k$ ensures position encodings to have the same dimension as the vectors of $Q$, $K$ and $V$.

To attend to different information from different representation subspaces jointly, the \emph{multi-head attention}~\cite{vaswani2017attention} strategy is applied as a conventional operation, where a head is an independent scaled dot-product attention module:
\begin{align}
\label{eq:multihead}
\begin{split}
    \text{MultiHead}(Q, K, V) &= \text{ConCat}(\text{head}_1,...,\text{head}_h)W_O, \\
    \text{head}_i &= \text{Attention}(QW_{Qi}, KW_{Ki}, VW_{Vi}),
\end{split}
\end{align}
where $W_{Qi}$, $W_{Ki}$, $W_{Vi}\in \mathbb{R}^{D\times d_{ki}}$ are the linear transformation parameters same as in Eq.~\eqref{eq:qkv} and $W_{O}$ are the linear transformation parameters for aggregating the extracted information from different heads.
Note that $d_{ki} = \frac{d_{k}}{h}$ and $d_{ki}$ must be an aliquot part of $d_{k}$. $h$ is the total number of the attention heads and we use two heads in the implementation.

Then the GApool is used to extract the temporal dependencies between steps by taking as input the output of the self-attention module and output an encoded representation.

The lower stream that exploits the dynamic interactions among agents works in the same way but the spatial dependencies among agents are encoded by the hidden states of an LSTM. Finally, the outputs of these two streams are connected and passed to a FC layer for fusion. The fused information includes dynamic spatial-temporal context.

\subsection{Multiple Trajectories Prediction}
\label{subsec:prediction}
Our method is CVAE-based and predicts multiple trajectories by repeatedly sampling from a learned latent space conditioned on the encoded information.
The CVAE is an extension of the VAE \cite{kingma2014auto} by introducing a condition to control the output \cite{sohn2015learning}.
Given a set of samples $(\mathbf{X, Y})=((\mathbf{X}_1, \mathbf{Y}_1 ),\cdots,(\mathbf{X}_m, \mathbf{Y}_m))$, it jointly learns a recognition model $q_\phi(\mathbf{z}|\mathbf{Y},\,\mathbf{X})$ of a variational approximation of the true posterior $p_\theta(\mathbf{z}|\mathbf{Y},\,\mathbf{X})$ and a generation model $p_\theta(\mathbf{Y}|\mathbf{X}, \,\mathbf{z})$ for predicting the output $\mathbf{Y}$ conditioned on the input $\mathbf{X}$. $\mathbf{z}$ are the stochastic latent variables, $\phi$ and $\theta$ are the respective recognition and generative parameters. The goal is to maximize the \textit{Conditional Log-Likelihood}:
  $ \log{p_\theta(\mathbf{Y}|\mathbf{X})} =  \log\sum_{\mathbf{z}}  p_\theta(\mathbf{Y}, \mathbf{z}|\mathbf{X})
    = \log{(\sum_{\mathbf{z}} q_\phi(\mathbf{z}|\mathbf{X}, \mathbf{Y})\frac{p_\theta(\mathbf{Y}|\mathbf{X}, \mathbf{z})p_\theta(\mathbf{z}|\mathbf{X})}{q_\phi(\mathbf{z}|\mathbf{X}, \mathbf{Y})})}.$
According to Jensen's inequality~\cite{jensen1906fonctions}, the evidence lower bound can be obtained:
\begin{equation}
\label{eq:CVAE}
\begin{split}
    \log{p_\theta(\mathbf{Y}|\mathbf{X}}) \geq &
    - D_{KL}(q_\phi(\mathbf{z}|\mathbf{X}, \,\mathbf{Y})||p_\theta(\mathbf{z}))
    + \\
    & \mathbb{E}_{q_\phi(\mathbf{z}|\mathbf{X}, \,\mathbf{Y})}
    [\log p_\theta(\mathbf{Y}|\mathbf{X}, \,\mathbf{z})],
\end{split}
\end{equation}
where $p_\theta(\mathbf{z})$ is made statistically independent from $p_\theta(\mathbf{z|\mathbf{X}})$~\cite{kingma2014semi,sohn2015learning}.
Here both the approximated posterior $q_\phi(\mathbf{z}|\mathbf{X}, \,\mathbf{Y})$ and the prior $p_\theta(\mathbf{z})$ are assumed to be Gaussian distribution for an analytical solution \cite{kingma2014auto}. During training, the Kullback-Leibler divergence $D_{KL}(\cdot)$ acts as a regularizer and pushes the approximated posterior to the prior distribution $p_\theta(\mathbf{z})$. The generation error $\mathbb{E}_{q_\phi(\mathbf{z}|\mathbf{X}, \,\mathbf{Y})}(\cdot)$ measures the distance between the generated output and the ground truth. During inference, for a given observation $\mathbf{X}_i$, one latent variable $\mathbf{z}_i$ is drawn from the prior distribution $p_\theta(\mathbf{z})$, and one of the possible output $\hat{\mathbf{Y}}_i$ is generated from the distribution $p_\theta(\mathbf{Y}_i|\mathbf{X}_i,\mathbf{z}_i)$. The latent variables $\mathbf{z}$ allow for the one-to-many mapping from the condition to the output via multiple sampling.
In this work, we model a conditional distribution $p_\theta(\mathbf{Y}_n|\mathbf{X})$, where $\mathbf{X}$ is the observed trajectory information and $\mathbf{Y}_n$ is one of its possible future trajectories. 

\textbf{Training:} As shown in Fig.~\ref{fig:pipeline}, during the training, both the observed trajectory $\mathbf{X}_i$ and its future trajectory $\mathbf{Y}_i$ are encoded by Encoder X and Y (see Sec.~\ref{subsec:encoder}), respectively. Then, their encodings are concatenated and passed through two FC layers (each is followed by a ReLU activation) for fusion. Then, two side-by-side FC layers are used to estimate the mean $\mu_{z_i}$ and the standard deviation $\sigma_{z_i}$ of the latent variables $\mathbf{z}_i$. 
A trajectory $\hat{\mathbf{Y}}_i$ is reconstructed by an LSTM decoder step by step by taking $\mathbf{z}_i$ and the encodings of observation as input. Because the random sampling process of $\mathbf{z}_i$ can not be back propagated during training, the standard reparameterization trick \cite{kingma2014auto} is adopted to make it differentiable. 
To minimize the error between the predicted trajectory $\hat{\mathbf{Y}}_i$ and the ground truth $\mathbf{Y}_i$, the reconstruction loss is defined as the $L2$ loss (Euclidean distance). Thus, the whole network is trained by minimizing the loss function using the stochastic gradient descent method:
\begin{equation}
L = \|\hat{\mathbf{Y}}-\mathbf{Y}\|^2 + D_{KL}(q_\phi(\mathbf{z}|\mathbf{X}, \,\mathbf{Y})||\mathcal{N}(0,I)).
\end{equation}

\textbf{Test:} In the test phase, the ground truth of future trajectory is no more available and its pathway is removed (color coded in green in Fig.~\ref{fig:pipeline}).
A latent variable $\mathbf{z}$ is sampled from the prior distribution $\mathcal{N}(0,I)$ and concatenated with the observation encodings that serve as the condition for the following trained decoder, so that the decoder can predict a trajectory.
To predict multiple trajectories, this process (sampling and decoding) is repeated multiple times. 

\subsection{Trajectory Ranking}
\label{subsec:ranking}
We propose a ranking strategy to select the \textit{most-likely} predicted trajectory out of the multiple predictions in order to adjust the Trajnet challenge setting. 
We apply bivariate Gaussian distribution to rank the predicted trajectories $(\hat{\mathbf{Y}}_{i,1},\cdots,\hat{\mathbf{Y}}_{i,N})$ for each agent. At step $t'$, all the predicted positions for agent $i$ are stored in $|{\hat{\mathsf{X}}_{i}}, {\hat{\mathsf{Y}}_{i}}|^{t'}$. 
We follow~\cite{graves2013generating} to fit the positions into the probability density function: 
\begin{equation}
\label{eq:ped}
\begin{split}
  & f(\hat{x}_i, \hat{y}_i)^{t'} = \frac{1}{2\pi\sigma_{\hat{\mathsf{X}}_{i}}\sigma_{\hat{\mathsf{Y}}_{i}}\sqrt{1-\rho^2}}\exp{\frac{-Z}{2(1-\rho^2)}}, \\
Z &= \frac{(\hat{x}_i-\mu_{\hat{\mathsf{X}}_{i}})^2}{{\sigma_{\hat{\mathsf{X}}_{i}}}^2} + \frac{(\hat{y}_i-\mu_{\hat{\mathsf{Y}}_{i}})^2}{{\sigma_{\hat{\mathsf{Y}}_{i}}}^2} -  \frac{2\rho(\hat{x}_i-\mu_{\hat{\mathsf{X}}_{i}})(\hat{y}_i-\mu_{\hat{\mathsf{Y}}_{i}})}{\sigma_{\hat{\mathsf{X}}_{i}}\sigma_{\hat{\mathsf{Y}}_{i}}}.
\end{split}
\end{equation}
where $\mu$ denotes the mean and $\sigma$ the standard deviation, and $\rho$ is the correlation between $\hat{\mathsf{X}}_{i}$ and $\hat{\mathsf{Y}}_{i}$.
A predicted trajectory is scored as the sum of the relative likelihood of all its steps:
  $  S(\mathbf{\hat{Y}}_{i,n}) = \sum_{t'=T+1}^{T'}f(\hat{x}_i, \hat{y}_i)^{t'}.$
All predicted trajectories are ranked by this score and the one with the highest score stands out for the single-path prediction.

\section{Experiments}
\label{sec:experiment}

To evaluate the performance of our proposed method, 
we compare DCENet with the most influential and recent nine state-of-the-art models from the Trajnet \cite{sadeghiankosaraju2018trajnet} challenge leader-board for a fair comparison: (1) \emph{Linear (off)}: a simple temporal linear regressor; (2) \emph{Social Force}~\cite{helbing1995social}: the very high impact rule-based model that implements social force to avoid collisions; (3)
\emph{S-LSTM}~\cite{alahi2016social}: the highly cited LSTM-based model that introduces social pooling layer for modeling interactions; (4)
\emph{S-GAN}~\cite{gupta2018social}: a GAN-based trajectory predictor; (5) \emph{MX-LSTM}~\cite{hasan2018mx}: an LSTM trajectory predictor that utilizes the head direction of agent;
(6) \emph{SR-LSTM}~\cite{zhang2019sr}: an LSTM-based model that refines the hidden states by message passing; 
(7) \emph{RED}~\cite{becker2018evaluation}: an RNN encoder-decoder model predicts trajectory only using observations; (8) \emph{Ind-TF}~\cite{giuliari2020transformer}: a Transformer-based trajectory predictor;
(9) \emph{AMENet}~\cite{cheng2021amenet}: the most recent state-of-the-art on the Trajnet leader-board.
We further design a series of ablation studies to analyze the impact of each proposed module, \ie~dynamic maps, transformer and LSTM encoder/decoder: (1) \emph{Baseline:} an LSTM encoder-decoder only using the observed trajectory as input;
(2) \emph{DCENet w/o DMs:} the stream of encoding dynamic maps is removed from our final model;
(3) \emph{Trans. En\&De:} the LSTM encoder-decoder is substituted by the Transformer encoder/decoder~\cite{vaswani2017attention} in our framework.

\subsection{Datasets}
\label{subsec:datasets}
\textbf{Trajnet}~\cite{sadeghiankosaraju2018trajnet} is one of the most popular forecasting benchmarks.
In Trajnet, 8 consecutive ground-truth locations (3.2 seconds) of each trajectory are for observation and the following 12 steps (4.8 seconds) are required to forecast.
Trajnet is a superset of diverse popular benchmark datasets: ETH~\cite{pellegrini2009you}, UCY~\cite{lerner2007crowds}, Stanford Drone Dataset~\cite{robicquet2016learning}, BIWI Hotel \cite{pellegrini2009you}, and MOT PETS~\cite{ferryman2009pets2009}. There is a total of 11448 trajectories from these four subsets covering 38 scenes for training. The test data is from the diverse partitions of them (besides MOT PETS) of the other 20 scenes without ground truth. The Trajnet challenge provides a specific server for online evaluation. 
It is worth noting that many existing works are evaluated on a subset of Trajnet using their own train/test splits.
For the sake of fairness, we only compare DCENet to the works which have shown their performance on the Trajnet challenge leader-board.

\begin{table}[t!]
\centering
\caption{
Results of different methods on the Trajnet challenge~\cite{sadeghiankosaraju2018trajnet}.
Models are categorized into deterministic (determ.) and stochastic (stoch.) depending on whether they incorporate a generative module.}
\setlength{\tabcolsep}{4pt}
\begin{tabular}{lllll}
\toprule
Model                & Category         & Avg. [m]$\downarrow$ & FDE [m]$\downarrow$   & ADE [m]$\downarrow$ \\
\midrule

S-LSTM~\cite{alahi2016social}         & determ.        & 1.3865            & 3.098 & 0.675  \\
S-GAN~\cite{gupta2018social}          & stoch.         & 1.3340            & 2.107 & 0.561  \\
MX-LSTM~\cite{hasan2018mx}            & determ.        & 0.8865            & 1.374 & 0.399  \\
Linear (off)                          & determ.        & 0.8185            & 1.266 & 0.371   \\
Social Force~\cite{helbing1995social} & determ.        & 0.8185            & 1.266 & 0.371  \\
SR-LSTM~\cite{zhang2019sr}            & determ.        & 0.8155            & 1.261 & 0.370  \\
RED~\cite{becker2018evaluation}       & determ.        & 0.7800            & 1.201 & 0.359  \\
Ind-TF~\cite{giuliari2020transformer} & determ.        & 0.7765            & 1.197 & 0.356  \\ 
AMENet~\cite{cheng2021amenet}         & stoch.         & 0.7695            & 1.183 & 0.356  \\ \midrule 
Baseline                              & stoch.         & 0.8045            & 1.239 & 0.370   \\
DCENet w/o DMs                        & stoch.         & 0.7760            & 1.195 & 0.357  \\
Trans. En\&De                         & stoch.         & 0.7780            & 1.196 & 0.360  \\ 
DCENet                                & stoch.         & \textbf{0.7660}   & \textbf{1.179} & \textbf{0.353} \\

\bottomrule
\end{tabular}
\label{tb:results}
\end{table}

\textbf{inD} was acquired by Bock \etal~\cite{inDdataset} using drones at four busy intersections in Germany in 2019. The traffic is dominated by vehicles and they interact with pedestrians heavily. 
The speed difference and confrontation makes the trajectory prediction challenging.
The data was processed to obtain the same format as Trajnet: 8 steps for observation and the following 12 steps for prediction.

\subsection{Evaluation Metrics}
\label{subsec:evaluationmetrics}
We adopt the most popular evaluation metrics: the mean average displacement error (ADE) and the final displacement error (FDE) to measure the trajectory prediction performance.
ADE measures the aligned Euclidean distance from the prediction to its corresponding ground truth trajectory averaged over all steps. The mean value across all the trajectories is reported. FDE measures the Euclidean distance between the last position from the prediction to the corresponding ground truth position. 
In addition, the most-likely prediction is decided by the ranking method as described in Sec~\ref{subsec:ranking}. Compared with the ground truth (only if it is available), $@top10$ is the one out of ten predicted trajectories that has the smallest ADE and FDE. 

The implementation details of training and testing our methods can be found in our \href{https://github.com/wtliao/DCENet}{code repository}.

\subsection{Results}
\label{subsec:results}

\begin{figure*}[t!]
    \centering
    \begin{subfigure}{0.25\textwidth}
    \includegraphics[trim=0cm 0cm 0cm 0cm, width=1\textwidth]{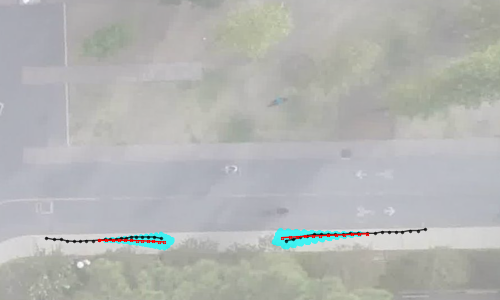}
    \label{subfig:bookstore-3}\vspace{-0.6cm}
    \caption{Trajnet bookstore-3}
    \end{subfigure}%
    \begin{subfigure}{0.25\textwidth}
    \includegraphics[trim=0cm 0cm 0cm 0cm, width=1\textwidth]{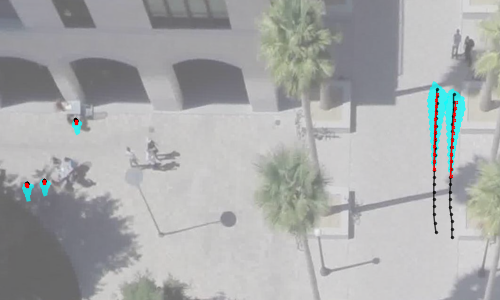}
    \label{subfig:coupa-3}\vspace{-0.6cm}
    \caption{Trajnet coupa-3}
    \end{subfigure}%
    \begin{subfigure}{0.25\textwidth}
    \includegraphics[trim=0cm 0cm 0cm 0cm, width=1\textwidth]{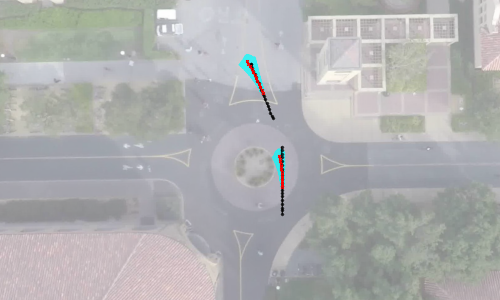}
    \label{subfig:deathCircle-0}\vspace{-0.6cm}
    \caption{Trajnet deathCircle-0}
    \end{subfigure}%
    \begin{subfigure}{0.25\textwidth}
    \includegraphics[trim=0cm 0cm 0cm 0cm, width=1\textwidth]{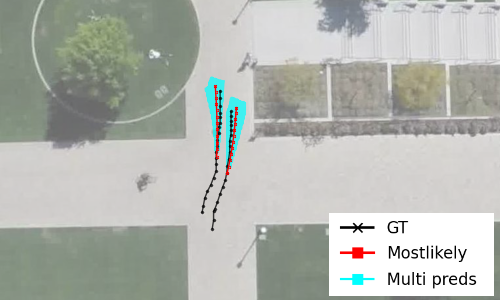}
    \label{subfig:hyang-6}\vspace{-0.6cm}
    \caption{Trajnet hyang-6}
    \end{subfigure}
    
    \begin{subfigure}{0.25\textwidth}
    \includegraphics[trim=0cm 0cm 0cm 0cm, width=1\textwidth]{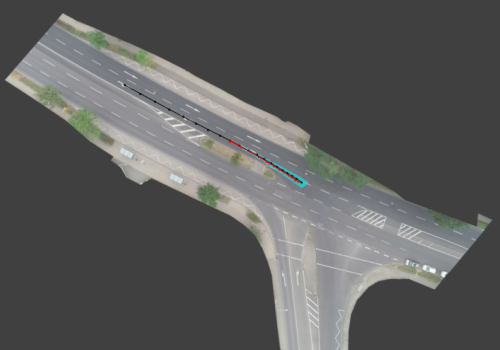}
    \label{subfig:intersectionA}\vspace{-0.6cm}
    \caption{inD Intersection-(A)}
    \end{subfigure}%
    \begin{subfigure}{0.25\textwidth}
    \includegraphics[trim=0cm 0cm 0cm 0cm, width=1\textwidth]{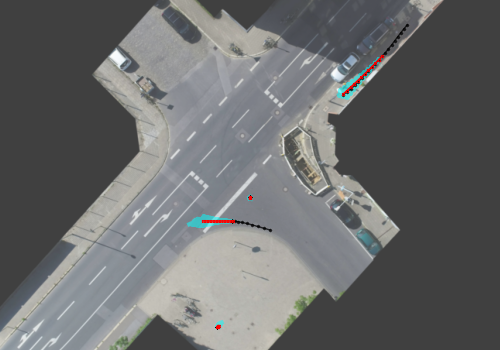}
    \label{subfig:intersectionB}\vspace{-0.6cm}
    \caption{inD Intersection-(B)}
    \end{subfigure}%
    \begin{subfigure}{0.25\textwidth}
    \includegraphics[trim=0cm 0cm 0cm 0cm, width=1\textwidth]{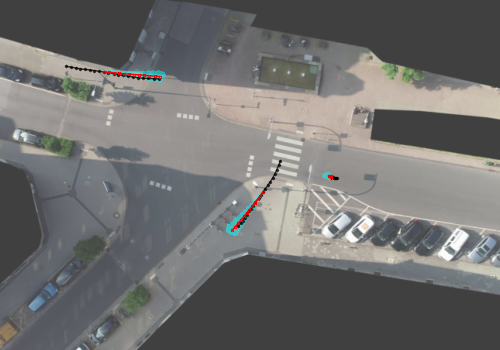}
    \label{subfig:intersectionC}\vspace{-0.6cm}
    \caption{inD Intersection-(C)}
    \end{subfigure}%
    \begin{subfigure}{0.25\textwidth}
    \includegraphics[trim=0cm 0cm 0cm 0cm, width=1\textwidth]{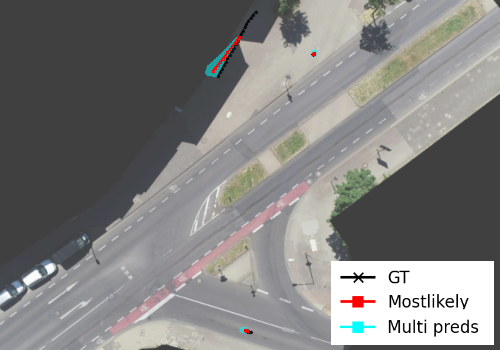}
    \label{subfig:intersectionD}\vspace{-0.6cm}
    \caption{inD Intersection-(D)}
    \end{subfigure}
    \caption{Multi-path trajectory predictions in shared spaces in Trajnet (1st row) and at different intersections in inD (2nd row). 
    }
    \label{fig:qualitative}
\end{figure*}

The experimental results from different methods including our ablative models reported on the Trajnet leader-board are listed in Table~\ref{tb:results}. Without ground truth trajectories, the single-path trajectory prediction was selected by the ranking mechanism.
We can see that DCENet reported new state-of-the-art performance and the ablative models also had comparable performances compared to the previous works.

\begin{table}[t!]
\centering
\caption{Quantitative results of our model and the comparative models on the inD benchmark measured by ADE/FDE.}
\begin{tabular}{lllll}
\toprule
Model                 & S-LSTM  & S-GAN      & AMENet   & DCENet \\ \midrule
inD             & \multicolumn{4}{c}{\textit{@top 10}}          \\ 
Intersection-(A)        & 2.04/4.61    & 2.84/4.91 & 0.95/1.94            & \textbf{0.72}/\textbf{1.50} \\
Intersection-(B)        & 1.21/2.99    & 1.47/3.04 & 0.59/1.29            & \textbf{0.50/1.07}          \\
Intersection-(C)        & 1.66/3.89    & 2.05/4.04 & 0.74/1.64            & \textbf{0.66/1.40}                   \\
Intersection-(D)        & 2.04/4.80    & 2.52/5.15 & 0.28/0.60            & \textbf{0.20}/\textbf{0.45} \\
Avg.            & 1.74/4.07    & 2.22/4.29 & 0.64/1.37            & \textbf{0.52/1.23}          \\ \midrule
inD             & \multicolumn{4}{c}{\textit{Most-likely}}                                      \\ 
Intersection-(A)        & 2.29/5.33    & 3.02/5.30 & 1.07/2.22            & \textbf{0.96}/\textbf{2.12} \\
Intersection-(B)        & 1.28/3.19    & 1.55/3.23 & 0.65/1.46            & \textbf{0.64/1.41}          \\
Intersection-(C)        & 1.78/4.24    & 2.22/4.45 & \textbf{0.83/1.87}   & 0.86/1.93                   \\
Intersection-(D)        & 2.17/5.11    & 2.71/5.64 & 0.37/0.80            & \textbf{0.28}/\textbf{0.62} \\
Avg.            & 1.88/4.47    & 2.38/4.66 & 0.73/1.59            & \textbf{0.69/1.52}          \\ \bottomrule
\end{tabular}
\label{tb:resultsinD}
\end{table}

First, by comparing to the Baseline, both DCENet w/o DMs and Ind-TF had much better results, and DCENet w/o DMs was slightly better in the average score and FDE but a little inferior in ADE than Ind-TF. 
Considering both models only use observed trajectories as input, it indicates that our method (self-attention + LSTM encoder/decoder) explored a better spatial-temporal context than Transformer. Furthermore, Ind-TF utilizes BERT, a heavily stacked Transformer structure and must be pre-trained on an external large-scale dataset, while DCENet does not require it. The results of DCENet w/o DMs indicates that its superior performance is not because we used more information (dynamic maps).

Second, by the comparison between the Baseline and S-LSTM, our Baseline model was significantly better. The difference between them is that our Baseline is CVAE-based and generates multiple trajectories. It indicates that the future motion of humans is of high uncertainty, and predicting a set of possible trajectories is better than only predicting a single one. It also demonstrates the effectiveness of the trajectory ranking methods (Sec.~\ref{subsec:ranking}), which was used to select the most-likely trajectory from the multiple predictions.
Our Baseline outperformed S-GAN significantly, which is a generative model for multiple trajectories prediction.

Third, interestingly, Trans. En\&De that adopts the Transformer encoder and decoder in our framework did not achieve improved performance compared to DCENet. This phenomenon indicates that our self-attention + LSTM encoder/decoder structure explored better dynamic context between agents than Transformer encoder/decoder in terms of trajectory prediction. The superior performance of DCENet w/o DMs against Ind-TF has also confirmed that.

Lastly,  DCENet outperformed DCENet w/o DMs. It indicates that the dynamic maps helped model the interactions between agents and were useful for trajectory prediction.

\textbf{Discussion} According to the comparison above, the results indicate: (1) DCENet is effective for predicting accurate trajectories for heterogeneous agents in various real-world traffic scenes, even without modeling interactions explicitly (the Baseline model). (2) The ranking method correctly estimates the multiple predictions and recommends a reliable candidate for the single-path trajectory prediction task. (3) Compared to the Baseline model, DCENet learns interaction via the dynamic maps with the self-attention structure effectively and shows improved performance. (4) Both LSTM and Transformer networks are capable of learning complex sequential patterns but their combination further enhances the performance in terms of trajectory prediction.

Furthermore, we have tested DCENet on inD~\cite{inDdataset} to justify its performance and generalization ability. We compare our model with the three most relevant models: S-LSTM for comparing with its occupancy grid mapping for agent-to-agent interaction, S-GAN for its generative module, and AMENet for its CVAE module and LSTM sequential modeling.
To guarantee a fair comparison, all the models were trained and tested using the same data. S-LSTM predicts the distributions of the positions~\cite{alahi2016social}. During inference, multiple positions were generated by sampling. Table~\ref{tb:resultsinD} lists the performance measured by ADE/FDE. Our model achieved the best performance for the $@top10$ prediction across all the intersections and reduced the errors by a big margin. Our model also outperformed the other models for the most-likely prediction at three out of four intersections. It only slightly fell behind the AMENet model on the intersection-(C). We anticipate that the most-likely prediction fell behind the $@top10$ prediction. However, the ranking method was still effective in recommending a reliable candidate in comparison to the other models. The results indicate: (1) Our model is able to generalize on different datasets and maintain superior performance. (2) Predicting multiple paths is more beneficial than predicting a single one for an agent. On the one hand, multiple predictions increase the chances to narrow down the errors. On the other hand, a single prediction may lead to a wrong conclusion especially if the initial steps predicted are deviating from the ground truth and the errors will accumulate significantly with time.
The multiple predictions form into an area indicating the potential intent of an agent and the area size reflects the uncertainty of an agent's intent.  

The qualitative results are shown in Fig.~\ref{fig:qualitative}. 
The first row showcases the scenarios in the Trajnet dataset.
Note that the qualitative analysis on Trajnet was carried out on the validation set (an independent subset of the training set) for comparing with the ground truth. Our model accurately predicted two pedestrians walking towards each other at bookstore-3. The shadow areas indicate multiple possible trajectories. 
It also correctly predicted the static pedestrians in coupa-3, as well as the pedestrians walking in parallel. In deathCircle-0, our model predicted different possible turning angles for the cyclist in the roundabout. 
In hyang-6, two pedestrians walking closely to each other were predicted correctly.
The second row showcases the scenarios in the inD dataset. Our model predicted a fast driving vehicle with a slightly different predicted speed at the Intersection-(A). It predicted that a left-turning vehicle may turn at the intersection-(B) with varying tuning angle and speed. 
The model also correctly predicted the interaction at the zebra crossing at the intersection-(C), where the vehicle stops to yield the way to the pedestrian. Similar predictions can be seen for the walking and static pedestrians, as well as the vehicle waiting at the entrance of the intersection-(D).
Overall, we can also see that the recommended single path is very close to the corresponding ground truth for each agent.

\section{Conclusion}
\label{sec:conclusion}

In this paper, we proposed a novel framework DCENet for multi-path trajectory prediction for heterogeneous agents in various real-world traffic scenarios. 
We decompose the learning of dynamic spatial-temporal context into exploiting the dynamic spatial context between agents using self-attention and the LSTM encoder and learning temporal context between steps with the following self-attention and global average pooling.
The spatial-temporal context is encoded into a latent space using a CVAE module. 
Finally, a set of future trajectories for each agent is predicted conditioned on the spatial-temporal context using the trained CVAE module. 
DCENet was evaluated on the Trajnet challenge benchmark and achieved the new state-of-the-art performance on the leader-board. 
Its superior performance on the inD benchmark further validated its efficacy and generalization ability. 
The ablation studies justified the impact of each module in DCENet. 
In the future, we are interested in extending the method for learning the impact from environment/static context, \eg~space layout and scene deployment, to further enhance the performance of trajectory prediction.

\pagebreak
\bibliographystyle{IEEEtran}
\bibliography{bibliography}

\end{document}